\documentclass[conference]{IEEEtran}
\IEEEoverridecommandlockouts
\usepackage{cite}
\usepackage{amsmath,amssymb,amsfonts}
\usepackage{algorithmic}
\usepackage{graphicx}
\usepackage{textcomp}
\usepackage{xcolor}

\usepackage{eso-pic}

\newcommand{\IEEEPostingNotice}{%
© 2026 IEEE.  Personal use of this material is permitted.  Permission from IEEE must be obtained for all other uses, in any current or future media, including reprinting/republishing this material for advertising or promotional purposes, creating new collective works, for resale or redistribution to servers or lists, or reuse of any copyrighted component of this work in other works.
}

\newcommand{\AIMArxivFirstPageNotice}{%
  \AddToShipoutPictureFG*{%
    \AtPageLowerLeft{%
      \raisebox{0.30in}{%
        \hspace*{\dimexpr 1in+\hoffset+\oddsidemargin\relax}%
        \parbox{\textwidth}{%
          \fontsize{5.5}{6.4}\selectfont
          \IEEEPostingNotice\\[-0.2ex]
          Published in the 2026 IEEE/ASME International Conference
          on Advanced Intelligent Mechatronics (AIM).
          DOI: 10.1109/AIM65483.2026.11658185.
        }%
      }%
    }%
  }%
}

\def\BibTeX{{\rm B\kern-.05em{\sc i\kern-.025em b}\kern-.08em
    T\kern-.1667em\lower.7ex\hbox{E}\kern-.125emX}}

\begin{document}

\title{\LARGE \bf
Multi-bounce Drum Roll with Optimized Active Tricks to Leverage Soft Embodiment
}


\author{Naoto Yamanaka$^{1}$, Takanori Jin$^{1}$ and Taisuke Kobayashi$^{1}$
\thanks{*This work was supported by JST, CREST, Japan Grant Number JPMJCR2555.}
\thanks{$^{1}$N. Yamanaka, T. Jin, and T. Kobayashi are with the National Institute of Informatics (NII) and with The Graduate University for Advanced Studies (SOKENDAI),
        2-1-2 Hitotsubashi, Chiyoda-ku, Tokyo, 101-8430, Japan
        {\tt\small naoto\_yamanaka@nii.ac.jp}}}%

\maketitle
\AIMArxivFirstPageNotice

\begin{abstract}

This paper presents a soft robotic drummer for accurate and efficient drum rolls.
High-frequency drum rolls require the ``multi-bounce technique,'' where a drumstick bounces multiple times with a single stroke.
In robotic reproduction of this technique, the body's elasticity is key, while the fine motion during the stroke is also crucial for maximizing the potential of that elasticity.
Therefore, we design two tricks: i) Tap-Pull (TP) trick to increase the number of rebounds by adding a pulling motion after impact; and ii) Micro-Pulse (MP) trick to keep the drumming volume by injecting small oscillations during the stroke.
Due to the nonlinear complexity of soft embodiment, both tricks are efficiently tuned using Bayesian optimization in a data-driven manner for accomplishing the respective objectives quantified.
We evaluated the optimized behaviors with soft and rigid end-effectors.
As a result, the soft TP achieved the highest bounce count (12.25 per stroke) with uniform intervals.
The soft MP suppressed the volume decay, yielding 6.8-times higher acoustic efficiency compared to the rigid MP.
These results indicate that the proposed tricks with the combination of elasticity and optimization can make robots play excellent drum rolls.

\end{abstract}


\section{Introduction}

The robotic automation of musical performance has been widely studied across various instruments: for example, Toyota’s violin-playing robot \cite{Kusuda2008}; a saxophone-playing robot capable of controlling fingers, lips, tongue, and lungs \cite{Solis2010}; and a piano-playing robotic hand that mimics the skeletal structure of the human hand \cite{Hughes2018}.
More recently, drum performance by humanoid robots has been demonstrated in simulation \cite{Shahid2025}.
Unlike keyboard and wind instruments, drums produce sound determined by impact conditions and provide the rhythmic foundation of the performance.
Consequently, precise strike phase accuracy, high-speed motion, and coordination across joints are required \cite{Shahid2025}.

Within such a drum performance, drum rolls would be the most challenging technique.
Human drummers switch among multiple stroke types to achieve diverse timbres and tempos.
In particular, they use the different body parts dominantly according to the tempo required: the wrist for slow tempos; and the fingers (how to grip drumsticks) for fast tempos.
The latter is the combination of active strokes and passive bounces with a continuous sound akin to white noise, known as a drum roll.
Its frequency by experts is over 30 Hz \cite{Hajian1997}.
Reproducing such high-speed operation in humanoid robots like \cite{Shahid2025} is difficult due to many hardware limitations, such as friction, backlash, sensing latency, and so on.
Therefore, research on percussion performance (particularly the drum roll) has focused on its principles to achieve acoustic targets, not to imitate human's complex behaviors, with simplified mechanisms \cite{Bretan2016}.

Several prior studies on robotic drum rolls have considered the importance of elasticity.
By appropriately adjusting stiffness parameters of a single joint, a passive second hit followed each active first hit \cite{Hajian1997,Kim2014}.
Subsequently, Karbasi et al. \cite{karbasi2022, karbasi2023exploring} developed a robotic drummer with a spring mechanism and investigated how the amplitude and frequency of the striking trajectory affect the passive second hit.
They found that the interval between the first and second bounces does not depend on the trajectory amplitude or frequency, but varies with the spring constant.
Moreover, passive bounce involves not only the number and interval of bounces.
Wakle et al. \cite{Wakle2024} evaluated high-speed repeated strikes using a dielectric elastomer actuator.
In that study, multiple passive bounces were observed due to the elastic actuator, and it was also reported that the sound amplitude decays across successive bounces.
This decay occurs because kinetic energy is dissipated on each bounce, reducing impact velocity and thus sound amplitude.
In this way, introducing appropriate softness is important for producing drum rolls with multiple passive bounces.
However, the degree of softness basically determines the number and interval of bounces, and the bounces in turn determine the sound amplitude.

While softness (especially elasticity) is thus important for drum rolls, the optimal design/control of the stiffness parameters for the best drum rolls cannot be analyzed due to high nonlinearity and uncertain factors (e.g., a drum and drumsticks used) \cite{torin2014nonlinear}.
To address this, Karbasi et al. \cite{Karbasi2021} proposed a learning method that updates joint stiffness parameters on each trial according to feedback from acoustic outcomes.
They finally achieved the desired bounce count and intervals.
The same research team \cite{Karbasi2024} also trained a robot with a spring mechanism of fixed intrinsic stiffness to perform drum patterns consistent with MIDI scores.
Using intrinsically motivated reinforcement learning \cite{Barto2004}, they showed that internal rewards reinforce double and triple strokes induced by the intrinsic stiffness, enabling the robot to actively exploit passive bounces.
At the same time, however, they indicated sample-efficiency constraints and difficulty of real-time processing as limitations for reinforcement learning.
In addition, because real-time analysis of acoustic data is difficult, they compute rewards by analyzing recorded audio after the performance.

Summarizing insights from the above prior research, the following two points are considered open issues.
\begin{enumerate}
    \item While moderate softness (elasticity) is important for drum rolls, it cannot shape the sound waveform between strokes into the desired one since it is passive phenomena for impact.
    \item While data-driven optimization of drum rolls is effective, the low sample efficiency becomes critical since sound simulation remains in the research stage \cite{wang2025sound}, necessitating real-world learning.
\end{enumerate}
In addition, the reinforcement learning adopted in the literature \cite{Karbasi2024} involves action determination through feedback to the faced state (auditory information in this case), but such real-time feedback control might be difficult for high-frequency motions like drum rolls.

To resolve these issues, we propose another directions for accomplishing drum rolls, namely adding fine feedforward motions during strokes.
That is, we assume that human's drum-roll behaviors cannot be described by a purely passive model, but includes small active motions that adjust the sound between the impact of the drumstick and the subsequent swing back up.
Even the small motions should resonate with a soft body, capable of inducing significant effects.
By tuning the motions using a sample-efficient learning method, we can harness that effects to produce the desired sound.

Then, we first design and fabricate a dual-arm robot with soft arm using a 3D printer that can induce passive bounces.
For fully use of this robot, we introduce the following two tricks.
The one is Tap-Pull (TP) trick, which pulls the drumstick toward the performer after the initial impact to prolong the bounce, thereby increasing the number of bounces while maintaining uniform inter-bounce intervals.
The other is Micro-Pulse (MP) trick, which applies small-amplitude oscillations after the initial impact to compensate sound amplitude, thereby mitigating the decay of post-first-hit amplitudes while maintaining uniform inter-bounce intervals.

These tricks are defined with multiple parameters, so by tuning them, we can obtain the desired behaviors.
Since the tricks are conducted in a feedforward manner, a black-box optimization, without designing the state space like reinforcement learning requires in the previous work \cite{Karbasi2024}, is suitable.
To apply it, we define objective functions based on the sound analysis for each episode (such as, the inter-bounce interval, the decay ratio of sound volume, and the number of bounces).
Bayesian optimization (BO) \cite{Shahriari2015} is employed as a sample-efficient black-box optimization.

With the proposed tricks, the performance of drum rolls is evaluated.
The two tricks were confirmed to alter the bounce behavior after impact, with the soft arm amplifying the range of change relative to their parameters more than the rigid one.
As a result, executing the optimized tricks with the dual-arm robot increased the number of bounces to over 12 and improved acoustic efficiency by 6.8 times.

\section{System setup}
\subsection{Dual-arm robot and soft arm}
\begin{figure}[tbp]
  \centering
  \includegraphics[width=0.96\linewidth]{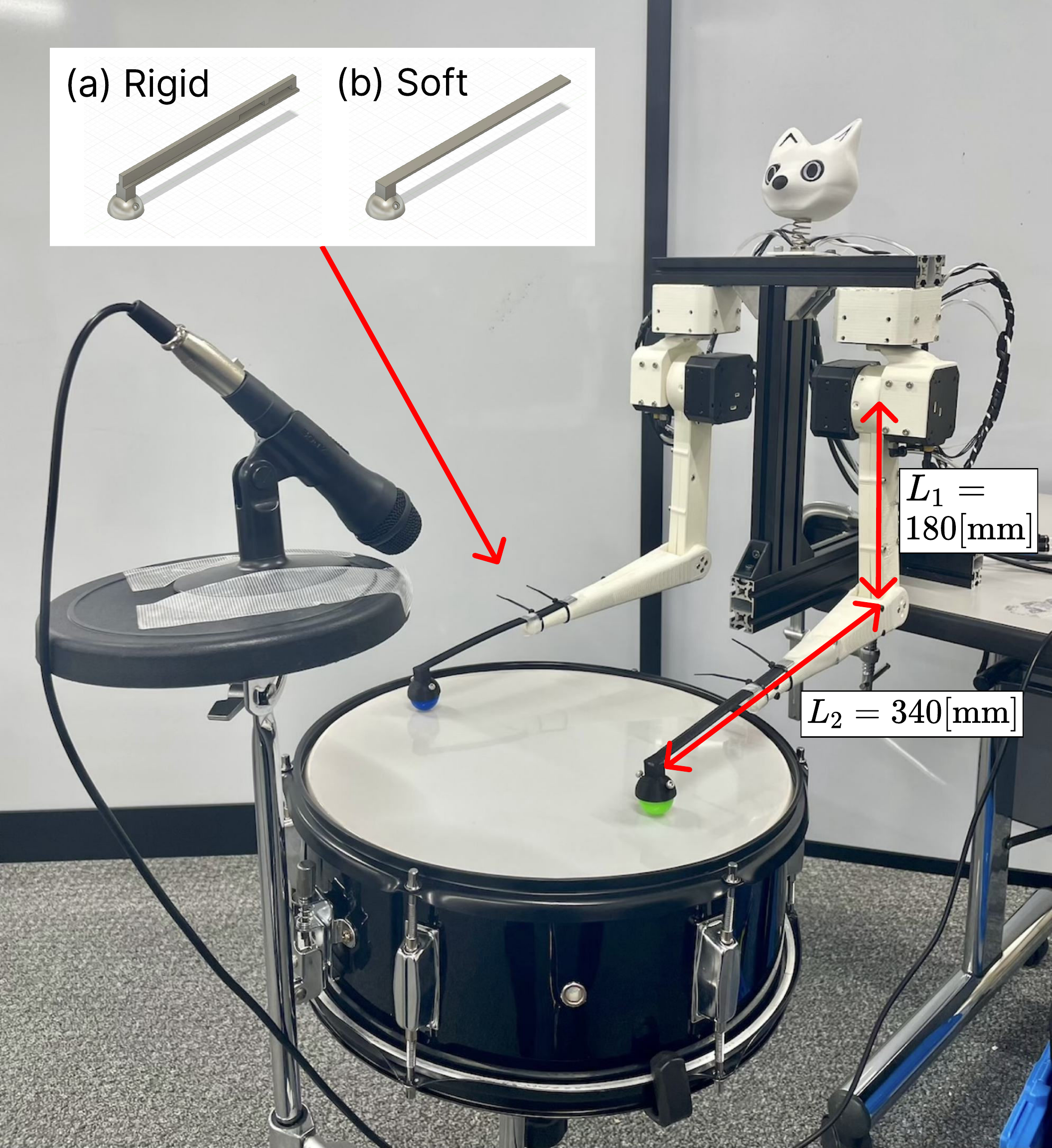}
  \caption{Experimental setup for drum-roll evaluation. The dual-arm robot strikes the snare drum while acoustic signals are recorded by the condenser microphone. Link lengths are $L_1=180$~mm and $L_2=340$~mm. Two end-effectors compared in this paper: (a) Rigid-arm (b) Soft-arm.}
  \label{fig:overview}
\end{figure}

Figure~\ref{fig:overview} shows the dual-arm robot designed for the experiments in this paper, each arm consisting of three joints.
The motors used are Koala BEAR V2 developed by Westwood Robotics.
In our experimental setup, joint-angle commands were updated at 200~Hz.
Because of their low gear ratio and water-cooling capability, they are suitable for repetitive tasks with collisions such as drum rolls.

As its end-effectors, soft and elastic arms made of PLA material with a thickness of $t=2$~mm were 3D-printed.
A rubber ball with strong elasticity is fixed by a screw at the tip to generate passive bounces upon impact.
To compare the effects of softness on bouncing behaviors, a rigid arm was also fabricated.
It adds vertical reinforcing ribs to the soft arm in order to prevent warping and bending, suppressing bouncing even when the tip rebounds.

\subsection{Processing of acoustic signals}
\label{subsec:signal}
A Sony condenser microphone (ECM-PCV80U) was used to sense the acoustic signals with a sampling rate of 44.1~kHz and a block size of 1024 samples.
The audio data were recorded as single-channel floating-point values within $[-1, 1]$ together with timestamp information for all samples.
After each episode, the following signal processing was applied to the recorded waveform.

First, the DC component was removed.
Then, a discrete Fourier transform was applied, and only the 1–100~Hz band was passed to remove high-frequency components and noise.
The band-passed signal was then processed by the Hilbert transform to obtain the envelope, which was further smoothed by a moving-average window of 0.02~s for stable extraction of amplitude variations.
Local maxima on the envelope were detected as impact peaks using the \texttt{signal.find\_peaks} function of SciPy.
The coefficient of variation, calculated as the ratio of the standard deviation to the mean of the inter-impact intervals, was used as an indicator of the stability of the bounce intervals.

\section{Active motion tricks}
\subsection{Desired drum rolls}
Brown et al. \cite{Brown2019} identified the uniformity between strokes as a key feature of a high-quality drum roll.
Uniformity can be considered in two aspects: the evenness of bounce intervals and the evenness of bounce amplitudes.
In addition, a white noise-like sound is preferable by the impacts as fast as possible.
Achieving this behavior by repeatedly swinging the arm is impractical both in terms of energy efficiency and operational speed, so more bounces within an efficient single swing are desired.
However, multiple bounces generated from a single strike gradually lose energy, causing the amplitude of each subsequent bounce to decrease \cite{Wakle2024}.
That is, maintaining uniform amplitude becomes more difficult as the number of bounces increases: in other words, they are in a trade-off relationship.

Therefore, we consider two types of drum rolls as follows:
\begin{enumerate}
    \item A maximal number of bounces with uniformity of bounce intervals
    \item Uniformity of sound amplitude and bounce intervals under the constant number of bounces
\end{enumerate}
Furthermore, since the latter can be satisfied by swinging the drumsticks at constant intervals, efficiency should also be considered as a prerequisite.
We design the respective tricks during each stroke that is expected to achieve the above two desired drum rolls, and optimize them according to the corresponding objective functions.

\subsection{Basic drum stroke}
The target end-effector positions of the dual-arm robot can be converted into the corresponding joint angles via inverse kinematics.
The basic trajectory is defined only along the vertical ($z$) axis of the drum surface as a step function.
The left and right arms alternately move up and down with period $T$.
The end-effector positions of the left and right arms at time $t$, $\mathbf{p}^\mathrm{L,R}(t)$ are given as:
\begin{align}
\mathbf{p}^\mathrm{L}(t) &= \begin{cases}
[0 \ 0 \ A_{\mathrm{U}}]^\top & (0 \leq t \bmod T < T / 2)
\\
[0 \ 0 \ A_{\mathrm{L}}]^\top & (T / 2 \leq t \bmod T < T)
\end{cases}
\\
\mathbf{p}^\mathrm{R}(t) &= \begin{cases}
[0 \ 0 \ A_{\mathrm{L}}]^\top & (0 \leq t \bmod T < T / 2)
\\
[0 \ 0 \ A_{\mathrm{U}}]^\top & (T / 2 \leq t \bmod T < T)
\end{cases}
\end{align}
Here, $A_\mathrm{U}$ and $A_\mathrm{L}$ represent the upper and lower positions, respectively.
Qualitatively, a larger $A_\mathrm{U} - A_\mathrm{L}$ would result in louder sound, while setting $A_\mathrm{L}$ smaller would increase impact but may affect the resonance of the sound and subsequent bounces.
This simple step function generates a basic alternating drum stroke where the two arms strike the drum in turn.
As a result, one impact occurs in every period $T$, and the striking frequency is $1/T$~Hz.
Based on this step-function control of drum strokes, two additional motion tricks are proposed in this section.

\subsection{Tap-Pull trick}
\begin{figure}[tbp]
  \centering
  \includegraphics[width=\linewidth]{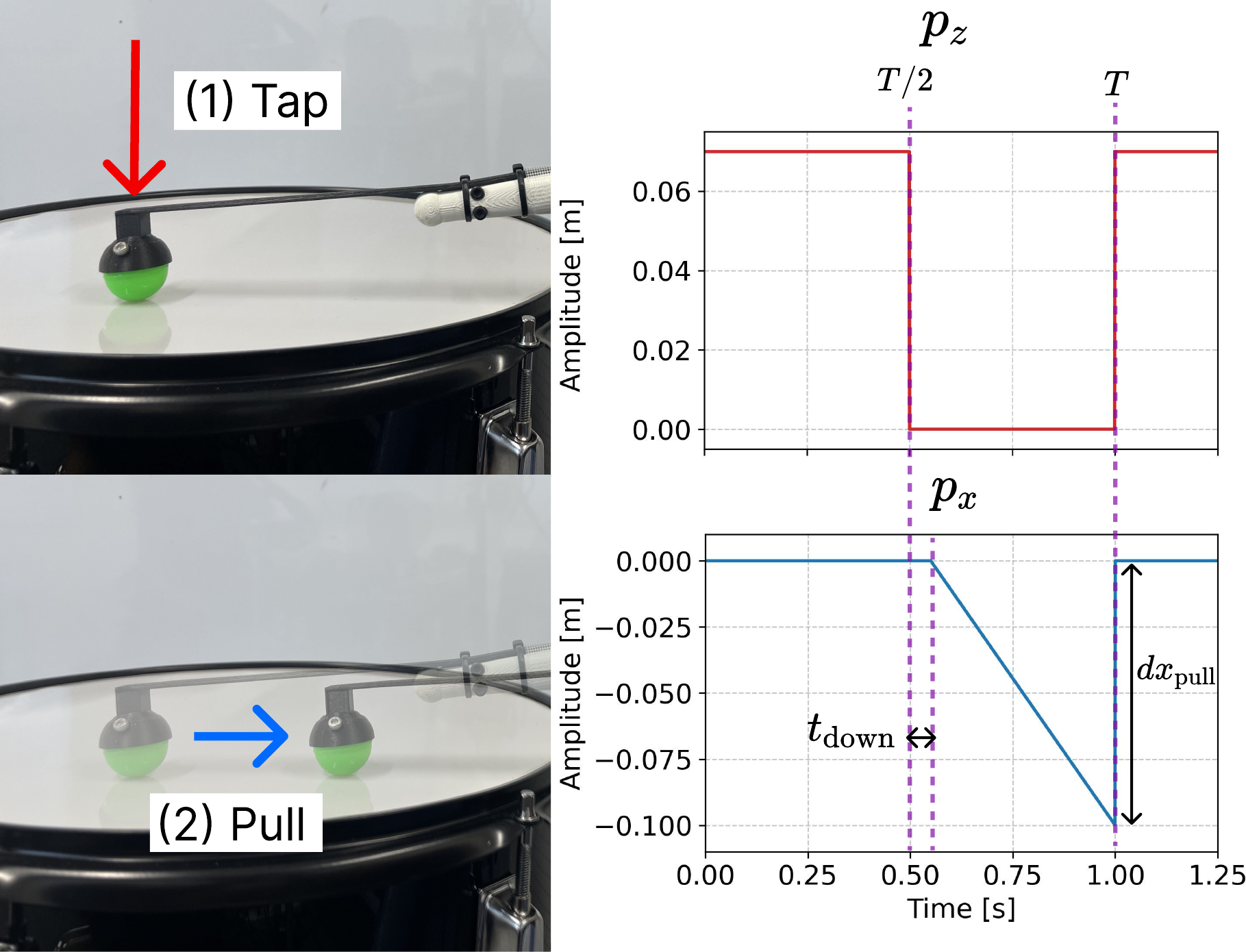}
  \caption{Tap-Pull (TP) trick and its command trajectory. The TP trick injects additional energy into passive rebounds by adding an active pull motion after a normal tap. Left: two phases of the motion (1) Tap (downward strike) and (2) Pull (horizontal retraction). Right: corresponding position commands: vertical command $p_z$ generates the tap, while horizontal command $p_x$ executes a linear pull of displacement $dx_\text{pull}$ after a waiting time $t_\text{down}$ during the downstroke.}
  \label{fig:tp_trick}
\end{figure}

Tap-Pull (TP) trick aims to achieve continuous sound by actively pulling the end effector toward the robot (Pull) after a normal strike (Tap). As illustrated in Fig.~\ref{fig:tp_trick}, TP consists of two phases:
(1) a normal downward strike (Tap) generated by the vertical step command $p_z$, and
(2) a horizontal retraction (Pull) implemented by a ramp command $p_x$.
The pull starts after a waiting time $t_\mathrm{down}$ (green dashed markers in Fig.~\ref{fig:tp_trick}) and ends with a total displacement $dx_\mathrm{pull}$.
This pulling motion injects additional kinetic energy necessary for bouncing, allowing the drum roll to continue without interruption.

Specifically, during $p^\mathrm{L,R}_z = A_\mathrm{L}$ with $T/2$ seconds, $p^\mathrm{L,R}_x$ linearly moves $dx_{\mathrm{pull}} (< 0)$ after waiting $t_{\mathrm{down}} (< T/2)$ seconds.
That is, given $t_{\mathrm{pull}} = T/2 - t_{\mathrm{down}}$, $p^\mathrm{L,R}_x$ at that time is updated as follows:
\begin{align}
    p^\mathrm{L,R}_x(t) = \frac{dx_{\mathrm{pull}}}{t_{\mathrm{pull}}} \max(0, (t \bmod T/2) - t_{\mathrm{down}})
\end{align}

\subsection{Micro-Pulse trick}
\begin{figure}[tbp]
  \centering
  \includegraphics[width=\linewidth]{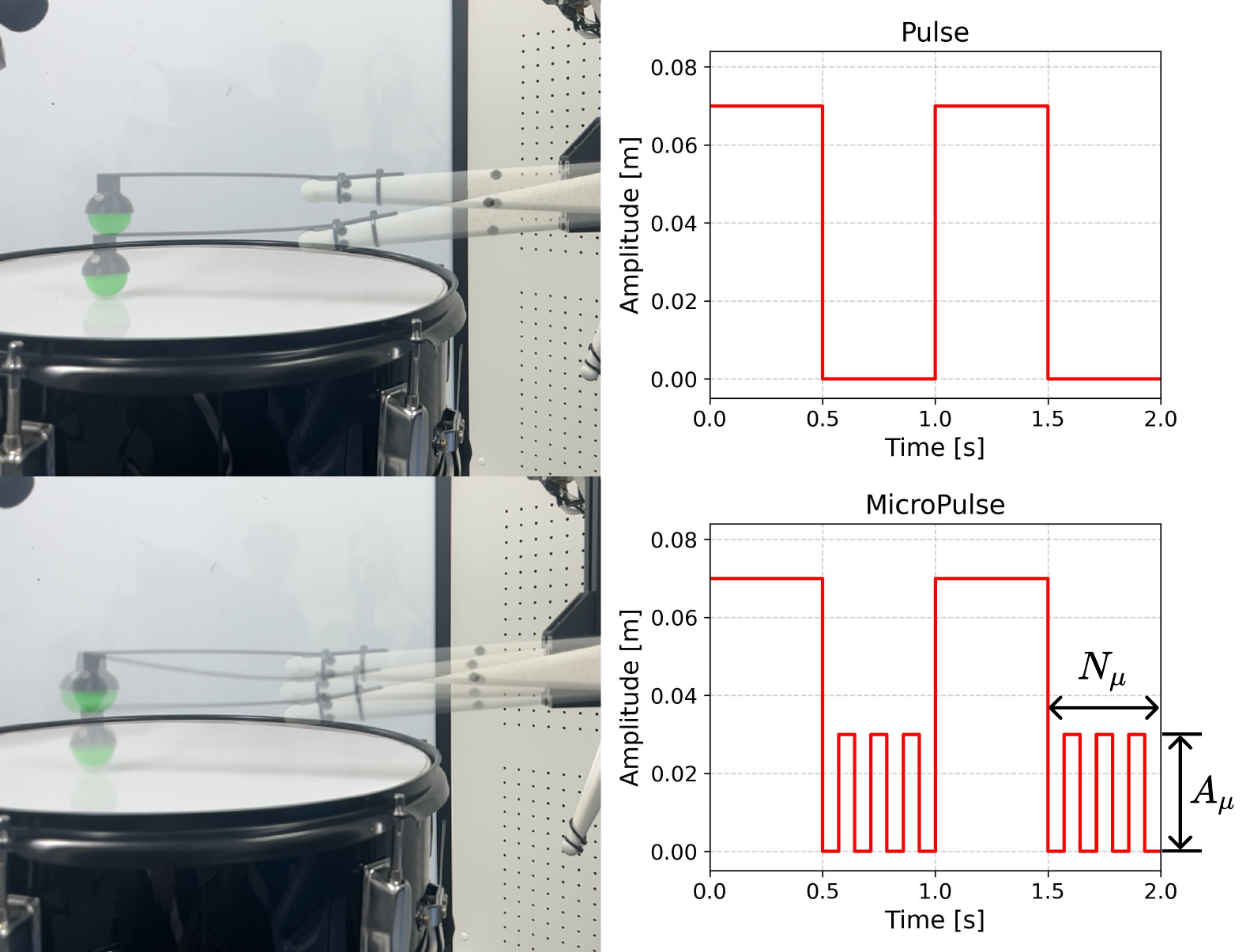}
  \caption{MP trick maintains drum-roll loudness by injecting small vertical oscillations during the downstroke to excite the elastic end-effector dynamics. Left: example stroke motion. Right: vertical command $p_z$ for a baseline pulse and the Micro pulse, where micro pulses of amplitude $A_\mu$ are inserted within the downstroke; $N_\mu$ controls the micro-pulse frequency (i.e., how finely the downstroke is segmented).}
  \label{fig:mp_trick}
\end{figure}

Micro-Pulse (MP) trick aims to improve the uniformity of sound amplitude by actively injecting fine pulses during each stroke to excite vibrations in the elastic arm.
Fig.~\ref{fig:mp_trick} compares the baseline command (Pulse) and the proposed MP command.
MP of amplitude $A_\mu$ are inserted $N_\mu$ times within each downstroke segment, thereby exciting the elastic dynamics without changing the overall stroke period.
This motion is expected to increase striking speed and thereby enhance the impact sound without decay.

Specifically, the duration of $p^\mathrm{L,R}_z = A_\mathrm{L}$ with $T/2$ seconds is divided into $2(N_{\mu}-1)+1$ segments.
Here, $N_{\mu}$ denotes the number of MP, which determines the frequency of the micro vibrations.
That is, given $\Delta t_m = (T/2) / (2(N_{\mu}-1)+1)$, $p^\mathrm{L,R}_z$ at that time is updated as follows:
\begin{align}
    p^\mathrm{L,R}_z(t) = \begin{cases}
        A_{\mathrm{L}} & 0 \leq (t \bmod T/2) \bmod \Delta t_m < \Delta t_m/2
        \\
        A_{\mu} & \Delta t_m/2 \leq (t \bmod T/2) \bmod \Delta t_m < \Delta t_m
\end{cases}
\end{align}
where, $A_{\mu} (> A_{\mathrm{L}})$ is the amplitude of the micro pulse.

\section{Optimization of tricks}
\subsection{Basic settings}
TP and MP tricks contain multiple parameters.
As the robot dynamics is nonlinear due to its soft embodiment, the effects of the parameters are difficult to analyze.
Therefore, this study automates parameter search in a data-driven manner.

Specifically, This process is time-consuming because it must be conducted in the real world due to the need of auditory information, the sample-efficient Bayesian optimization \cite{Shahriari2015} with the Tree-Structured Parzen Estimator \cite{Bergstra2011} as sampler, which is implemented in Optuna \cite{Akiba2019}, is employed.
The parameters for TP and MP tricks are optimized independently.
The respective parameters to be optimized and the corresponding objective functions to be minimized are described in the next sections.

\subsection{Optimization for TP trick}

\begin{table}[tbp]
\centering
\caption{TP: Control variables and search ranges}
\label{tab:tp_vars}
\begin{tabular}{lll}
\hline
Variable & Symbol & Search range\\
\hline
Upper amplitude [m] & $A_\mathrm{U}$ & [0.03, 0.07]\\
Lower amplitude [m] & $A_\mathrm{L}$ & [0.00, 0.02]\\
Downward timing [s] & $t_{\mathrm{down}}$ & [0.01, 0.05]\\
Pull distance [m] & $dx_{\mathrm{pull}}$ & [$-$0.10, 0.00]\\
\hline
\end{tabular}
\end{table}

The four parameters in TP trick (see Table~\ref{tab:tp_vars}) are optimized to minimize the following objectives.
\begin{align}
  \mathcal{L}^\mathrm{TP} = -N + \lambda_{\mathrm{TP}}\gamma_{\mathrm{int}}
  \label{eq:tp_obj}
\end{align}
where $N$ is the bounce count, $\gamma_{\mathrm{int}}$ is the uniformity metric of bounce intervals, and $\lambda_{\mathrm{TP}}>0$ is a weighting coefficient (set to $\lambda_{\mathrm{TP}}=5$ in this study).
The weight in Eq.~\eqref{eq:tp_obj} was manually chosen to balance the term scales and fixed across all conditions.
The objective encourages a larger number of bounces while improving the uniformity of bounce intervals.

Specifically, the bounce count $N$ is easily evaluated as the number of peaks detected in the acoustic data during one stroke (see the details in Section~\ref{subsec:signal}).
The time between bounces, $\Delta t_k$ ($k=1,\ldots,N-1$), is calculated using the bounce times $t_k$ as follows:
\begin{align}
\Delta t_k = t_{k+1}-t_k
\end{align}
The uniformity of bounce intervals, $\gamma_{\mathrm{int}}$, is then defined as the coefficient of variation:
\begin{align}
\gamma_{\mathrm{int}}
= \frac{\sigma(\{\Delta t_k\}_{k=1}^{N-1})}
       {m(\{\Delta t_k\}_{k=1}^{N-1})}
\end{align}
where
\begin{align*}
    m(\{x_k\}_{k=1}^{N-1}) &= \frac{1}{N-1}\sum_{k=1}^{N-1} x_k
    \\
    \sigma(\{x_k\}_{k=1}^{N-1}) &= \sqrt{\frac{1}{N-2}\sum_{k=1}^{N-1} (x_k - m(\{x_k\}_{k=1}^{N-1}))^2}
\end{align*}
If the bounce intervals are uniform, $\gamma_{\mathrm{int}}$ will be small.

\subsection{Optimization for MP trick}

\begin{table}[tbp]
\centering
\caption{MP: Control variables and search ranges}
\label{tab:mp_vars}
\begin{tabular}{lll}
\hline
Variable & Symbol & Search range\\
\hline
Micro amplitude [m] & $A_{\mu}$ & [0.00, 0.04]\\
Number of MP & $N_{\mu}$ & $\{1,\dots,7\}$\\
\hline
\end{tabular}
\end{table}

The two parameters in MP trick (see Table~\ref{tab:mp_vars}) are optimized to minimize the following objectives.
\begin{align}
  \mathcal{L}^\mathrm{MP} = \rho_{\mathrm{decay}} + \lambda_\gamma \gamma_{\mathrm{int}} + \lambda_N |N - N_{\mathrm{tgt}}| + \lambda_P P_M
  \label{eq:mp_obj}
\end{align}
where $\rho_{\mathrm{decay}}$ is the amplitude decay ratio, $\gamma_{\mathrm{int}}$ is the uniformity of bounce intervals, $N$ is the bounce count, $N_{\mathrm{tgt}}$ is the target bounce count, $P_{M}$ is the motor power, and $\lambda_{\gamma},\lambda_{N},\lambda_{P}>0$ are weights (set to $\lambda_{\gamma}=0.5$, $\lambda_{N}=0.2$, and $\lambda_{P}=8.0$ in this study).
The weights in Eq.~\eqref{eq:mp_obj} were manually chosen to balance the term scales and fixed across all conditions.
Note that the upper and lower amplitudes, $A_{U,L}$, are fixed at $0.04$~m and $0$~m, respectively, since the initial kinetic energy must be aligned for evaluating efficiency.
The desired drum roll, which should be acquired by the optimal MP trick, is to efficiently suppress the amplitude decay after the first strike while keeping the uniformity of bounce intervals.

Specifically, to fairly evaluate the efficiency and to keep the tempo, the number of bounces, $N$, needs to be controlled to a target bounce count $N_{\mathrm{tgt}}$.
As well as TP trick, the uniformity of bounce intervals, $\gamma_{\mathrm{int}}$, is also added to ensure the temporal regularity.
In addition, the amplitude of each impact is defined as $A_k$ ($k=1,\ldots,N$).
With this, the amplitude decay ratio, $\rho_{\mathrm{decay}}$, is given by
\begin{align}
  \rho_{\mathrm{decay}} = \frac{1}{N-1}\sum_{i=2}^{N}\left| 1 - \frac{A_k}{A_1} \right|
\end{align}
where $A_1$ is the special amplitude of the first impact.
$\rho_{\mathrm{decay}}$ should be zero to stabilize the sound by injecting the minimum necessary energy.

Finally, we consider the efficiency more directly.
Experienced human drummers achieve the required loudness with minimal muscle co-contraction by utilizing stick rebound \cite{Scott2020}.
Analogously, in this study, the ratio of acoustic power to motor power indicates acoustic efficiency, which should be maximized for louder sound with less motor effort.
To compute this ratio, the acoustic power, $P_A$, and motor power, $P_M$, are computed as follows:
\begin{align}
P_A &= \frac{4 \pi r^2 \kappa^2}{\rho_0 c} \frac{1}{K}\sum_{k=0}^{K-1} (s_k)^2
\\
P_M &= \frac{1}{K}\sum_{k=0}^{K-1} \tau_k\,\omega_k
\end{align}
where the signals are uniformly sampled with period $\Delta t$ into $K$ samples over the roll duration $T$, and $s_k$, $\tau_k$, and $\omega_k$ denote the sampled microphone signal, joint torque, and joint angular velocity at $\tilde{t}_k=k\Delta t$, respectively.
These parameters $r$, $\rho_0$, $\kappa$, and $c$ are acoustic-related constants and are difficult to identify accurately.
Therefore, obtaining a reliable numerical estimate of $P_A/P_M$ is not straightforward.
Hence, we consider the following as an alternative efficiency measure.
\begin{align}
    \eta = \frac{\hat{P}_A}{P_M}
\end{align}
where
\begin{align*}
    \hat{P}_A = \frac{1}{K}\sum_{k=0}^{K-1} (s_k)^2
\end{align*}
This is useful for relative evaluation between methods.
However, the maximization of $\eta$ (with the maximization of $\hat{P}_A$ and the minimization of $P_M$) would conflict with the other criteria since the maximization of $\hat{P}_A$ yields the maximization of $N$ and minimization of $\rho_{\mathrm{decay}}$.
Hence, only the minimization of $P_M$ is added to the objective function.

\section{Experiments}
\subsection{Setup}
We verify whether the proposed TP and MP tricks can achieve the desired drum rolls.
To achieve this, we compare four conditions involving i) the choice of soft or rigid arms and ii) the use of each trick.
Optimal parameters might vary by hardware, so the cases with different arms and tricks are sufficiently optimized individually via Bayesian optimization (200 iterations for TP trick with many parameters and 100 iterations for MP trick).
Note that parameters used regardless of trick (i.e. $A_\mathrm{U,L}$) reuse optimization results from trick-enabled to trick-disabled runs.
Subsequently, each condition is tested 10 trials using the acquired best parameters, with performance statistically evaluated via mean and standard deviation.
Evaluation metrics are the terms of each objective function; for the MP trick, $\hat{P}_A$ and $\eta$ are additionally computed.

\subsection{TP trick}
\begin{figure}[tbp]
  \centering
  \includegraphics[width=\linewidth]{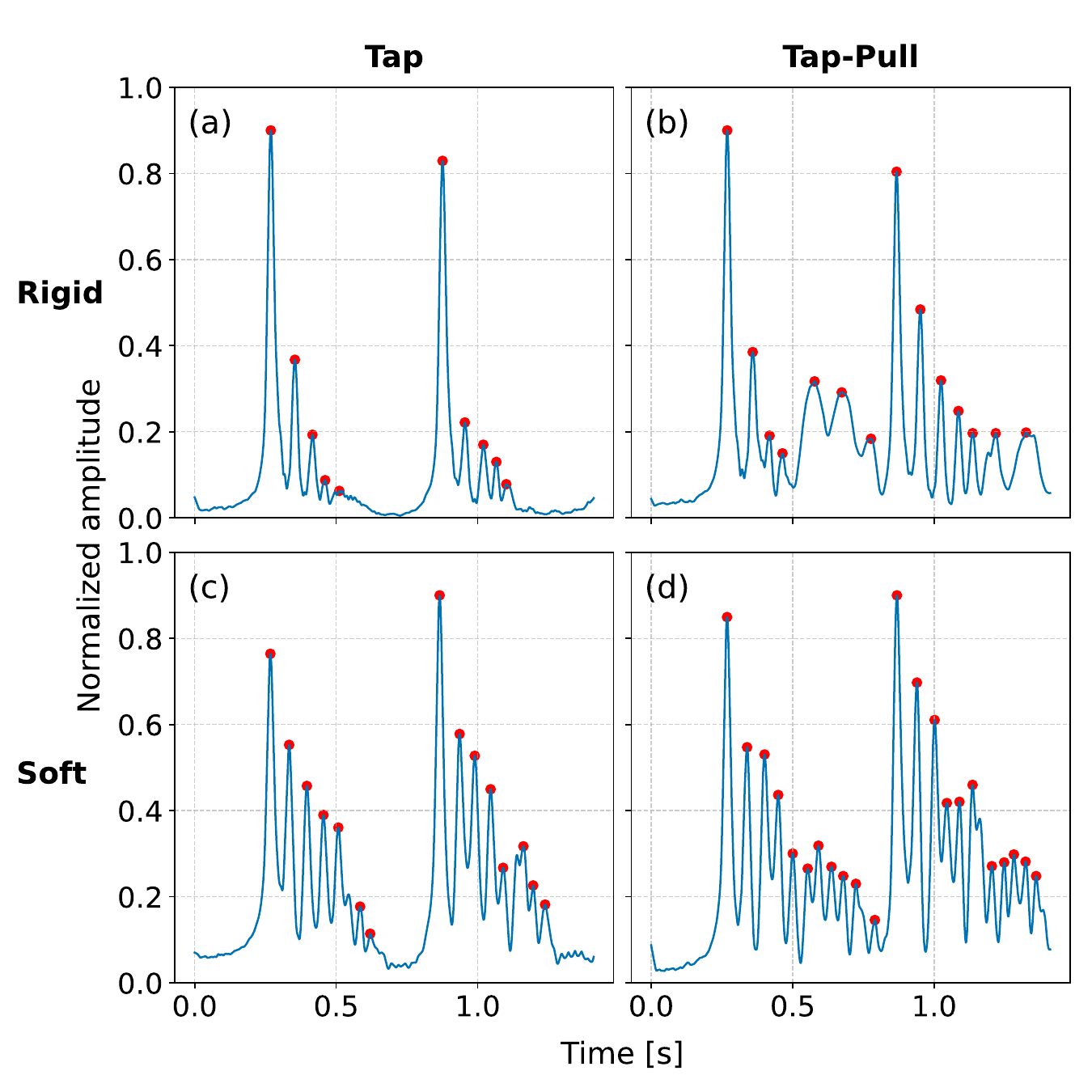}
  \caption{Peak detection for Tap and Tap–Pull (TP) strokes. The blue curves show the normalized amplitude envelope of the recorded drumming sound, and red markers indicate the detected peaks corresponding to impact events (including passive rebounds). (a) Rigid (Tap), (b) Rigid (Tap-Pull), (c) Soft (Tap), (d) Soft (Tap-Pull).}
  \label{fig:tappull_audio}
\end{figure}

The optimized parameters are summarized in Table~\ref{tab:tp_vars_opt}.
Table~\ref{tab:tappull_eval} lists the test scores, and Fig.~\ref{fig:tappull_audio} depicts the example of acoustic data at the tests.
Note that during trials, four strokes were executed in total by the left and right arms to confirm that the drum roll continued continuously.

Introducing TP trick increased the bounce count $N$ for both the rigid and soft arms, while the combination of soft arm and TP trick obtained the best, 12.25.
Also, it outperformed the others in terms of $\gamma_{\mathrm{int}}$, the uniformity of bounce intervals.
As a result, the optimized TP trick with the soft arm achieved the best drum rolls.

By comparing Fig.~\ref{fig:tappull_audio}a and Fig.~\ref{fig:tappull_audio}b, we can find instability of bounces when applying TP trick to the rigid arm.
With the rigid arm, the elasticity of the rubber ball attached to the tip causes slight bouncing after contact with the drum membrane, but since it is not flexible, the rebound is not natural and the resulting acoustics are more like dragging.
On the other hand, Fig.~\ref{fig:tappull_audio}d shows more stable waveform.
This difference also appears in $\gamma_{\mathrm{int}}$, and the soft arm showed the most uniform bounce intervals.

In addition, as can be seen in Fig.~\ref{fig:tappull_audio}c, even with the soft arm, the bouncing ended before swinging up the arm.
This caused time without sound between strokes, that is not desirable as drum rolls.
On the other hand, TP trick enabled continuous sound even between strokes (see Fig.~\ref{fig:tappull_audio}d).

\begin{table}[tbp]
\centering
\caption{TP: Results of Bayesian optimization}
\label{tab:tp_vars_opt}
\begin{tabular}{lllll}
\hline
Method& $A_\mathrm{U}$ & $A_\mathrm{L}$ & $t_{\mathrm{down}}$ & $dx_{\mathrm{pull}}$\\
\hline
Rigid(Tap) & 0.068 & 0.001 & - & -\\
Rigid(Tap-Pull) & 0.068 & 0.001 & 0.043 & -0.065\\
Soft(Tap) & 0.030 & 0.001 & - & -\\
Soft(Tap-Pull) & 0.030 & 0.001 & 0.029 & -0.055\\
\hline
\end{tabular}
\end{table}

\begin{table}[tbp]
\centering
\caption{Comparison of Tap and Tap-Pull. Values are means (standard deviations in parentheses).}
\label{tab:tappull_eval}
\begin{tabular}{lccc}
\hline
Method & Bounce/stroke & $\gamma_{\mathrm{int}}$ & $\mathcal{L}^\mathrm{TP}$ \\
\hline
Rigid, Tap      & $5.450(0.197)$ & $0.876(0.036)$ & $-1.069$ \\
Rigid, Tap-Pull & $7.125(0.358)$ & $0.364(0.036)$ & $-5.307$ \\
Soft, Tap       & $8.100(0.293)$ & $0.668(0.063)$ & $-4.756$ \\
Soft, Tap-Pull  & $\mathbf{12.25}(0.406)$ & $\mathbf{0.290}(0.053)$ & $\mathbf{-10.80}$ \\
\hline
\end{tabular}
\end{table}

\subsection{MP trick}
\begin{figure}[tbp]
  \centering
  \includegraphics[width=\linewidth]{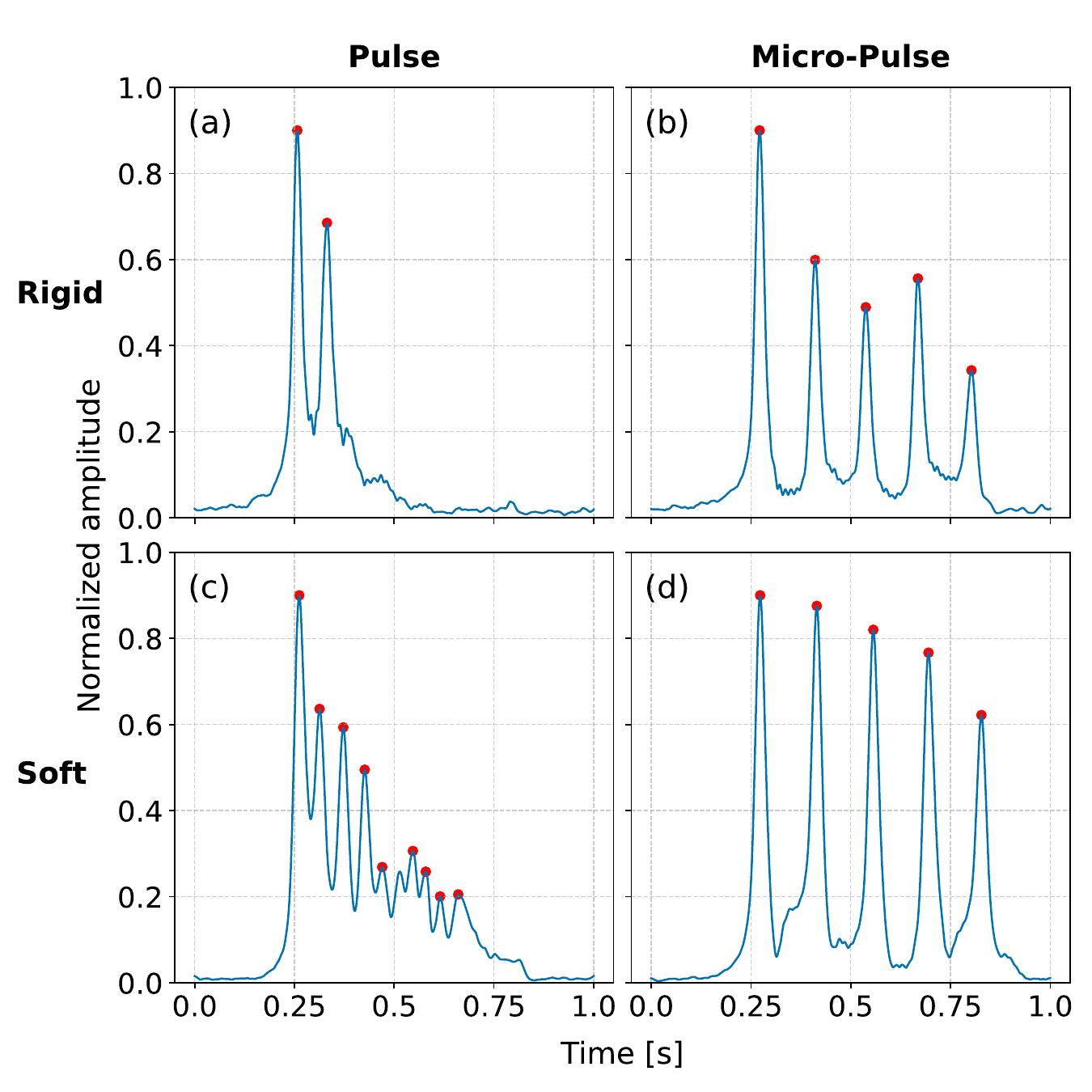}
  \caption{Peak detection examples for Pulse and Micro-Pulse (MP) strokes. The blue curves show the normalized amplitude envelope of the recorded drumming sound, and red markers indicate the detected peaks corresponding to impact events (including passive rebounds). (a) Rigid (Pulse), (b) Rigid (Pulse+Micro-Pulse), (c) Soft (Pulse), (d) Soft (Pulse+Micro-Pulse).}
  \label{fig:micro_audio}
\end{figure}

The optimized parameters are summarized in Table~\ref{tab:mp_vars_opt}.
Table~\ref{tab:micro_eval} lists the test scores, and Fig.~\ref{fig:micro_audio} depicts the example of acoustic data at the tests.
Note that during trials, only one stroke was played with a single arm.

For both the rigid and soft arms, the optimized MP trick successfully kept the number of bounces to the specified five times, and improved the uniformity of bounce intervals as well as the results for TP trick.
While the rigid arm could not reduce $\rho_{\mathrm{decay}}$ sufficiently even with the optimized MP trick, the soft arm with MP trick significantly reduced $\rho_{\mathrm{decay}}$.
These improvements are clearly apparent from Fig.~\ref{fig:micro_audio}.

However, the motor power $P_M$ of the soft arm with MP trick increased compared to the others.
This is natural because the dissipated energy needs to be injected by motors.
Therefore, by comparing the acoustic efficiency, $\eta$, we can say that the soft arm is better than the rigid arm (about 683\% improvement).
Note that the best $\eta$ was by the soft arm without MP trick, but this is due to excessive bounce count.
That is, the most efficient and well-organized drum rolls can be obtained by the case with the soft arm and the optimized MP trick.

\begin{table}[tbp]
\centering
\caption{MP: Results of Bayesian optimization}
\label{tab:mp_vars_opt}
\begin{tabular}{lll}
\hline
Method & $A_{\mu}$ & $N_{\mu}$\\
\hline
Rigid (Pulse) & - & - \\
Rigid (+Micro-Pulse) & 0.028 & 5 \\
Soft (Pulse) & - & -\\
Soft (+Micro-Pulse) & 0.032 & 5 \\
\hline
\end{tabular}
\end{table}

\begin{table*}[tbp]
\centering
\caption{Comparison of Pulse and Pulse+Micro-Pulse. Values are means (standard deviations in parentheses).}
\label{tab:micro_eval}
\begin{tabular}{lccccccc}
\hline
Method & $\rho_{\mathrm{decay}}$ & $P_M$ [W] & $\hat{P}_A$ & $\eta$ & Bounce & $\gamma_{\mathrm{int}}$ & $\mathcal{L}^\mathrm{MP}$ \\
\hline
Rigid (Pulse)        & $0.569(0.118)$ & $\mathbf{0.010}(0.001)$ & $0.001(0.000)$ & $0.133(0.000)$ & $3.800(0.632)$ & $0.569(0.118)$ & $2.178$ \\
Rigid (+Micro-Pulse) & $0.312(0.055)$ & $0.045(0.001)$ & $0.003(0.000)$ & $0.059(0.006)$ & $\mathbf{5.000}(0.000)$ & $\mathbf{0.025}(0.008)$ & $0.694$ \\
Soft (Pulse)         & $0.539(0.026)$ & $0.011(0.001)$ & $0.005(0.000)$ & $\mathbf{0.454}(0.006)$ & $9.200(0.422)$ & $0.225(0.076)$ & $4.212$ \\
Soft (+Micro-Pulse)  & $\mathbf{0.143}(0.001)$ & $0.058(0.001)$ & $\mathbf{0.023}(0.001)$ & $0.403(0.012)$ & $\mathbf{5.000}(0.000)$ & $0.029(0.004)$ & $\mathbf{0.636}$ \\
\hline
\end{tabular}
\end{table*}

\section{Discussion}
\begin{figure}[tbp]
  \centering
  \includegraphics[width=\linewidth]{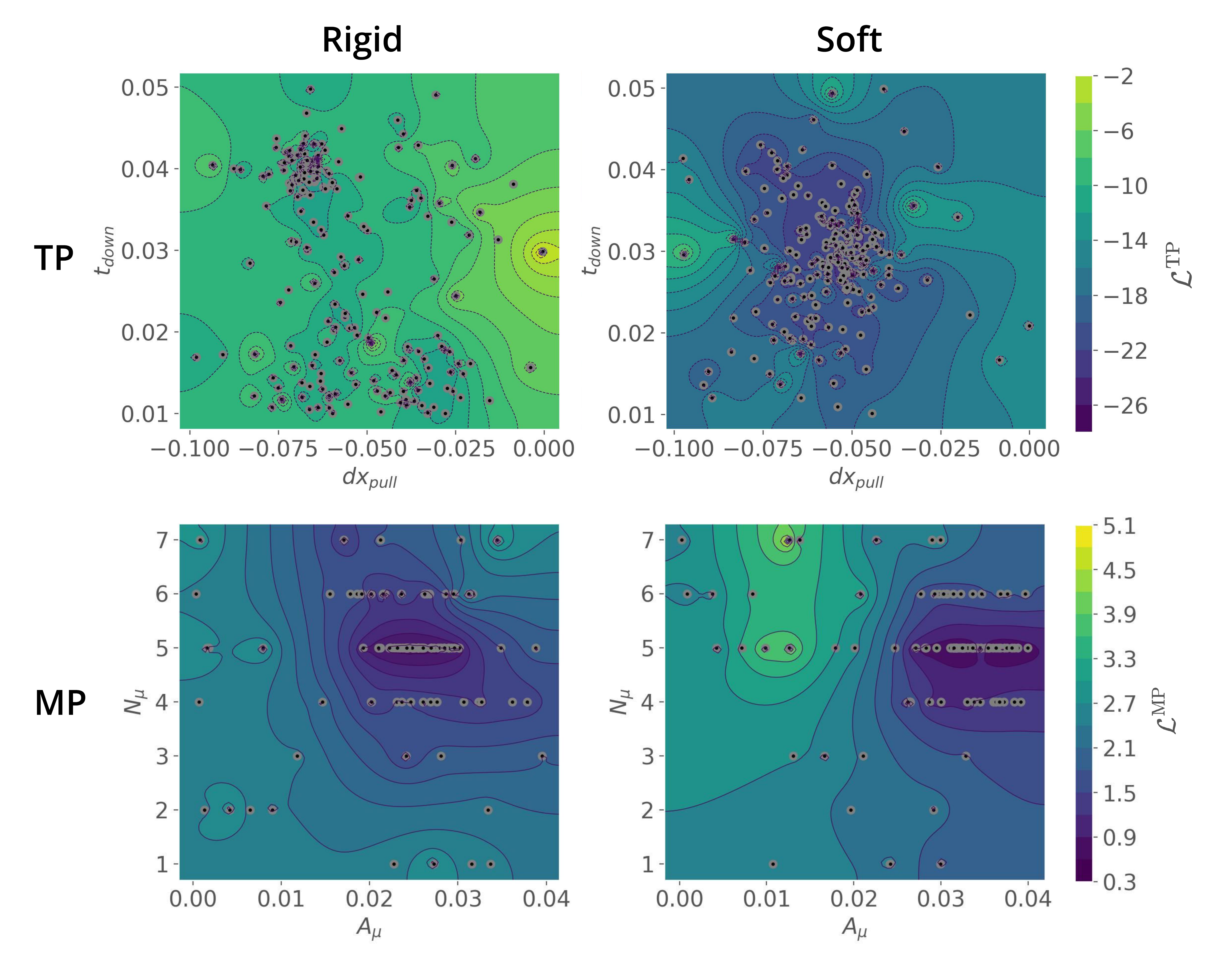}
  \caption{Heatmap of objective function estimated by Bayesian optimization for TP and MP tricks (rigid vs. soft).
Top row (TP): estimated objective value $\mathcal{L}^\mathrm{TP}$ over the parameter space ($dx_\text{pull}$, $t_\text{down}$). Bottom row (MP): estimated objective value $\mathcal{L}^\mathrm{MP}$ over ($A_\mu$, $N_\mu$). Left column shows the rigid end-effector, and right column shows the soft end-effector. }
  \label{fig:heatmap}
\end{figure}

We visualize the objective values for the selected parameters, which were estimated by Bayesian optimization in Fig.~\ref{fig:heatmap}.
Looking at these figures, the parameters for TP and MP tricks in the rigid arm do not contribute significantly to performance, suggesting that hardware performance can be exploited with relatively simple manual tunings.
In contrast, performance in the soft arm is highly sensitive to the changes of parameters, making it difficult to manually find optimal parameters.
This is precisely why optimization/learning techniques like the Bayesian optimization implemented in this paper should be important for fully exerting the potential of soft embodiment.

Nevertheless, several limitations should be noted.
First, the experiments were conducted under a fixed drum configuration and limited tempo range, and the generality of the observed parameter sensitivity across different physical setups remains to be verified.
Second, the objective function was purely quantitative, and perceptual aspects of drum roll quality were not explicitly evaluated.
Therefore, the relationship between the optimized objective values and human perceptual evaluation requires further investigation.

\section{Conclusion}
This study proposed two active motion tricks, TP and MP tricks, for a dual-arm drumming robot equipped with soft and elastic end-effectors, in order to actively inject energy into passive bounces.
Each trick was optimized according to its respective objective.
As a result, the proposed tricks successfully exerted the potential of soft embodiment, leading to the desired drum rolls with continuous sustained sound and uniformity and efficiency of sound amplitude.

As a possible future direction, the framework could be extended to multi-objective optimization, sound amplitude and bounce interval uniformity, energy efficiency, and the maximization of bounce count to achieve more desirable drum rolls expressivity.


\bibliographystyle{IEEEtran}
\bibliography{references.bib}

\end{document}